\documentclass[10pt,twocolumn,letterpaper]{article}
\pdfoutput=1

\usepackage{cvpr}
\usepackage{times}
\usepackage{epsfig}
\usepackage{graphicx}
\usepackage{amsmath}
\usepackage{amssymb}
\usepackage{color}
\usepackage{microtype}
\usepackage{multirow}

\usepackage{placeins}
\usepackage{xspace}
\usepackage{authblk}

\newcommand{\name}{Deep Feature Interpolation}
\newcommand{\nameshort}{DFI}

\DeclareMathOperator*{\argmin}{arg\,min}

\newcommand{\bx}{\mathbf{x}}

\newcommand{\bz}{\mathbf{z}}

\newcommand{\bw}{\mathbf{w}}

\newcommand{\source}{\ensuremath{\mathcal{S}^{s}}\xspace}
\newcommand{\target}{\ensuremath{\mathcal{S}^{t}}\xspace}
\newcommand{\intermediate}{\ensuremath{\phi(\bz)}\xspace}

\newcommand{\meansource}{\ensuremath{\bar{\phi}^{s}}\xspace}
\newcommand{\meantarget}{\ensuremath{\bar{\phi}^{t}}\xspace}

\newenvironment{packed_enum}{
\begin{enumerate}
  \setlength{\itemsep}{1pt}
  \setlength{\parskip}{2pt}
  \setlength{\parsep}{0pt}
}{\end{enumerate}}

\newif\ifcomments
\commentsfalse

\ifcomments
  \newcommand{\comments}[1]{#1}
\else
  \newcommand{\comments}[1]{}
\fi

\usepackage[pagebackref=true,breaklinks=true,letterpaper=true,colorlinks,bookmarks=false]{hyperref}

\cvprfinalcopy 

\def\cvprPaperID{3299} 

\ifcvprfinal\fi
\begin{document}

\title{\vspace{-1.4cm}Deep Feature Interpolation for Image Content Changes}


\renewcommand\Authsep{ }
\renewcommand\Authand{ }
\renewcommand\Authands{ }
\author[1,*]{\vspace{-1mm}Paul Upchurch}
\author[1,*]{\hspace{2.5mm}Jacob Gardner}
\author[1]{\hspace{2.5mm}Geoff Pleiss}
\author[2]{\hspace{2.5mm}Robert Pless}
\author[1]{\hspace{2.5mm}Noah Snavely}
\author[1]{\hspace{2.5mm}Kavita Bala}
\author[1]{Kilian Weinberger}
\affil[1]{Cornell University}
\affil[2]{George Washington University}
\affil[*]{Authors contributed equally}

\maketitle

\global\csname @topnum\endcsname 0
\global\csname @botnum\endcsname 0

\begin{abstract}
We propose {\em \name{} (\nameshort{})}, a new data-driven baseline
for automatic high-resolution image transformation. As the name
suggests, \nameshort{} relies only on simple linear interpolation of
deep convolutional features from pre-trained convnets. We show that
despite its simplicity, \nameshort{} can perform high-level semantic
transformations like ``make older/younger'', ``make bespectacled'',
``add smile'', among others, surprisingly well---sometimes even matching
or outperforming the state-of-the-art. This is particularly unexpected
as \nameshort{} requires no specialized network architecture or even any
deep network to be trained for these tasks. \nameshort{} therefore can be
used as a new baseline to evaluate more complex algorithms and provides
a practical answer to the question of which image transformation tasks
are still challenging after the advent of deep learning.

\end{abstract}

\begin{figure}[]
\centerline{
   \includegraphics[width=0.7\linewidth]{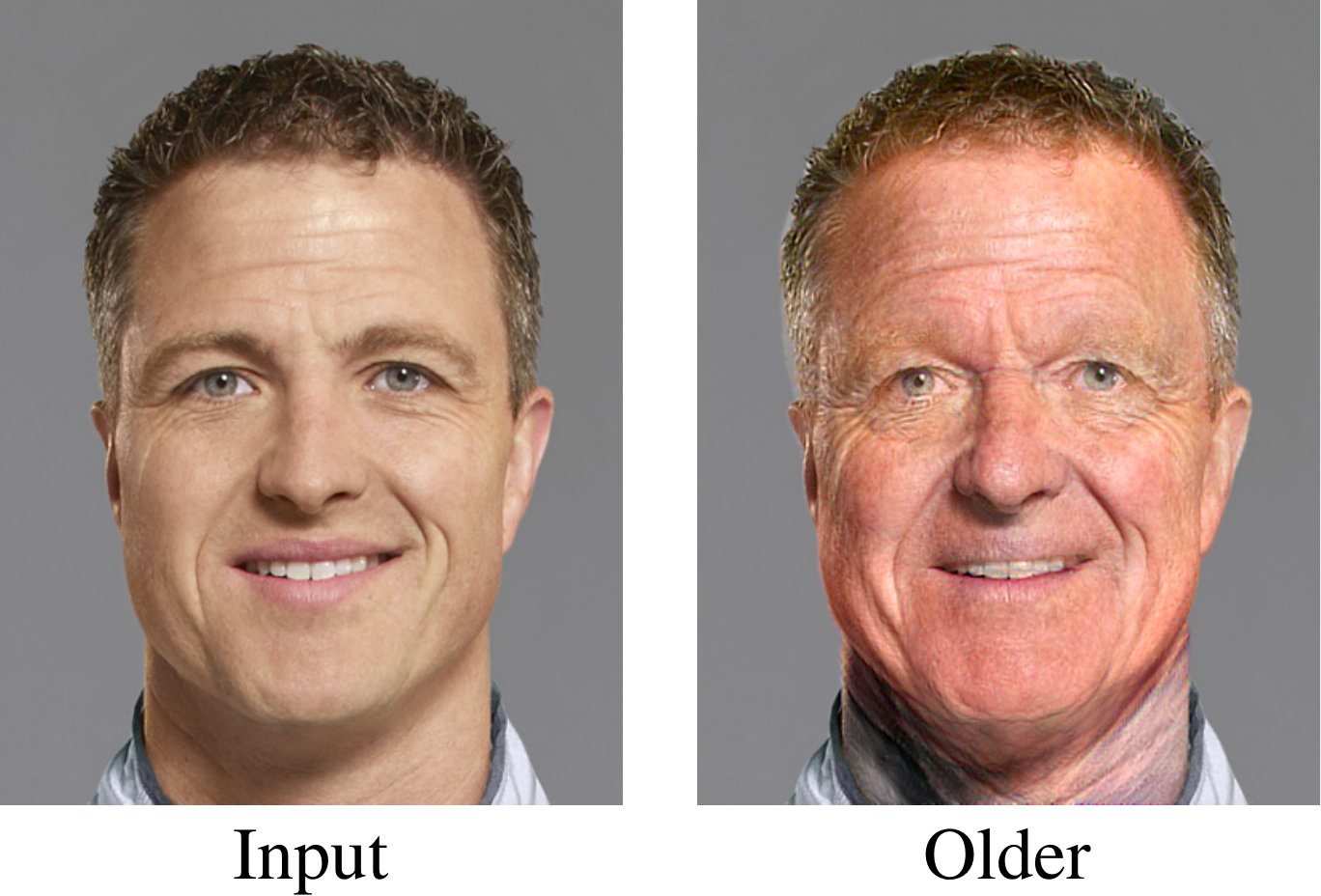}}
   \caption{Aging a face with \nameshort{}.}
\label{fig:teaser}
\end{figure}

\begin{figure*}[t]
\centerline{
   \includegraphics[width=0.7\linewidth]{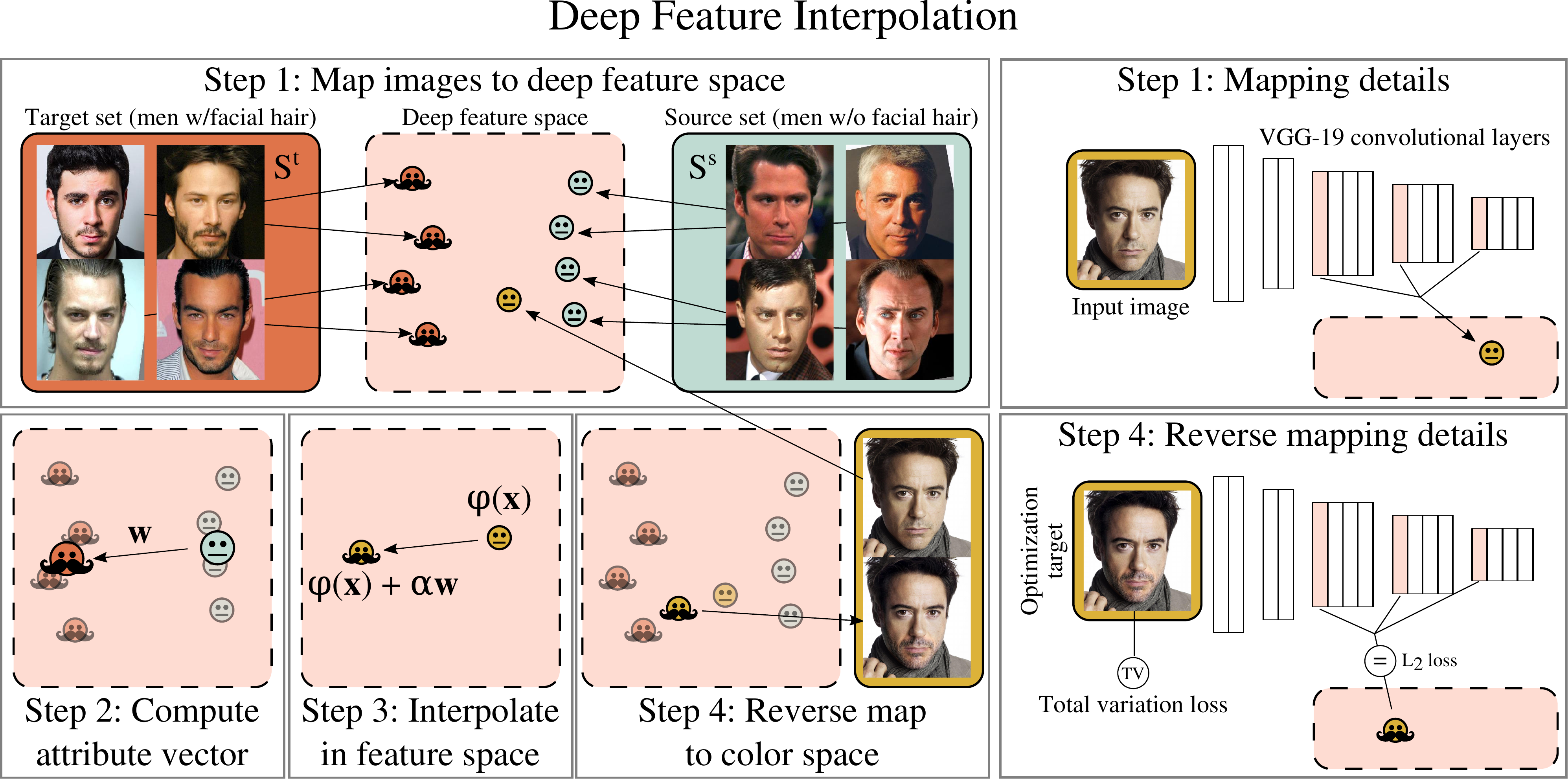}}
   \caption{A schematic outline of the four high-level \nameshort{} steps.}
\label{fig:overview}
\end{figure*}

\section{Introduction}

Generating believable changes in images is an active and challenging
research area in computer vision and graphics. Until
recently,   algorithms were typically hand-designed for individual transformation
tasks and exploited task-specific expert knowledge.  Examples include
transformations of human faces~\cite{suwajanakorn2015makes,kemelmacher2016transfiguring},
materials~\cite{bellini2016time,aittala2016reflectance}, color~\cite{zhang2016colorful}, or seasons in outdoor images~\cite{laffont2014transient}. However,
recent innovations in deep convolutional
auto-encoders~\cite{reed2014learning}  have produced a succession of more versatile
approaches.  Instead of  designing each algorithm for a specific task, a
conditional (or adversarial)
generator~\cite{kingma2014semi,goodfellow2014generative} can be trained to produce a set of possible
image transformations through supervised learning~\cite{yan2015attribute2image,wang2016generative,zhou2016view}.
Although these approaches can perform a variety of seemingly impressive tasks, in this paper we show that a surprisingly large set of them can be solved via  linear interpolation in deep feature space and may not require specialized deep architectures.

How can linear interpolation be effective? In pixel space, natural images lie on an (approximate) non-linear
manifold~\cite{weinberger2006unsupervised}. Non-linear manifolds are
locally Euclidean, but globally curved and non-Euclidean. It is well known that
in pixel space linear interpolation between images introduces ghosting
artifacts, a sign of departure from the underlying manifold, and linear
classifiers between image categories perform poorly.

On the other hand, deep convolutional neural networks (convnets) are known to
excel at classification tasks such as visual object
categorization~\cite{simonyan2014very,he2016deep,huang2016densely}---while relying on a simple linear layer at the end of the network for classification. These linear classifiers perform well
because networks  map images into new representations in which image classes
are \emph{linearly} separable. In fact, previous work has shown that neural
networks that are trained on sufficiently diverse object recognition classes, such as
VGG~\cite{simonyan2014very} trained on ImageNet~\cite{krizhevsky2012imagenet}, learn surprisingly
versatile feature spaces and can be used to train linear classifiers for additional image classes.   Bengio \etal~\cite{bengio2013better}
hypothesize that convnets linearize the manifold of natural images
into a (globally) Euclidean subspace of deep features.

Inspired by this hypothesis, we argue that, in such deep feature
spaces, some image editing tasks may no longer be as challenging as
previously believed. We propose a simple framework that leverages the notion that in the right feature space, image editing can be performed simply by linearly interpolating between images with a certain attribute and images without it.
For instance, consider the task of adding facial hair to the image of
a male face, given two sets of images: one set with facial hair, and one set without. 
If convnets can be trained to distinguish between male faces with
facial hair and those without, we know that these classes must be linearly separable. Therefore,
motion along a single linear vector should suffice to move an image from deep features corresponding
to ``no facial hair'' to those corresponding to ``facial hair". Indeed, we will show that even a simple choice of this vector suffices:
we average convolutional layer features of each set of images and take the difference. 

We call this method \name{} (\nameshort{}). Figure~\ref{fig:teaser} shows an example of a facial transformation with \nameshort{} on a $390 \times 504$ image.

Of course, \nameshort{} has limitations:  our method
works best when all images are aligned, and thus is suited
when there are feature points to line up (e.g. eyes and mouths in face images). It also requires that the sample images with and without the desired attribute are otherwise similar to the target image (e.g. in the case of Figure~\ref{fig:overview}, the other images should contain Caucasian males).

However, these assumptions on the data are comparable to what is typically used to train generative models, and in the presence of such data \nameshort{} works surprisingly well.
We demonstrate its efficacy on several transformation tasks commonly
used to evaluate generative approaches. 
Compared to prior work, it is much simpler, and often faster and more versatile: It does not require re-training a convnet,  is not specialized on any
particular task, and it is able to process much higher resolution images. 
Despite its
simplicity we show that on many of these image editing tasks it
outperforms state-of-the-art methods that are substantially more involved and
specialized.

\section{Related Work}

Probably the generative methods most similar to ours are~\cite{larsen2015autoencoding} and~\cite{radford2016unsupervised}, which similarly generate data-driven attribute transformations using deep feature spaces. We use these methods as our primary points of comparison; however, they rely on specially trained generative auto-encoders and are fundamentally different from our approach to learning image transformations. Works by Reed \etal~\cite{reed2014learning,reed2015deep} propose content change models for challenging tasks (identity and viewpoint changes) but do not demonstrate photo-realistic results.
A contemporaneous work~\cite{brock2016neural} edits image content by manipulating latent space variables. However, this approach is limited by the output resolution of the underlying generative model.
An advantage of our approach is that it works with pre-trained networks and has the ability to run on much higher resolution images. In general, many other uses of generative networks are distinct from our problem setting~\cite{goodfellow2014generative,denton2015deep,zhao2016energy,salimans2016improved,odena2016conditional,donahue2016adversarial,dumoulin2016adversarially}, as they deal primarily with generating novel images rather than changing existing ones.

Gardner \etal~\cite{gardner2015deep} edits images by minimizing the witness function of the Maximum Mean Discrepancy statistic. The memory needed to calculate the transformed image's features by their method grows linearly whereas \nameshort{} removes this bottleneck.

Mahendran and Vedaldi~\cite{mahendran15understanding} recovered visual imagery by inverting deep convolutional feature representations.
Gatys \etal~\cite{gatys2015neural} demonstrated how to transfer the
artistic style of famous artists to natural images by optimizing for
feature targets during reconstruction.
Rather than
reconstructing imagery or transferring style, we edit the content of an existing image while seeking to preserve photo-realism and all content unrelated to the editing operation.

Many works have used vector operations on a learned generative latent space to demonstrate transformative effects~\cite{dosovitskiy2015learning,radford2016unsupervised,girdhar2016learning,wu2016learning}. In contrast, we suggest that vector operations on a discriminatively-trained feature space can achieve similar effects.

In concept, our work is similar to~\cite{suwajanakorn2015makes,garrido2014automatic,thies2015real,kemelmacher2011exploring,kemelmacher2016transfiguring} that use video or photo collections to transfer the personality and character of one person's face to a different person (a form of puppetry~\cite{sumner2004deformation,weise2009face,kholgade2011content}). This difficult problem requires a complex pipeline to achieve high quality results. For example, Suwajanakorn \etal~\cite{suwajanakorn2015makes} combine several vision methods: fiducial point detection~\cite{xiong2013supervised}, 3D face reconstruction~\cite{suwajanakorn2014total} and optical flow~\cite{kemelmacher2012collection}. Our method is less complicated and applicable to other domains (e.g., product images of shoes).

While we do not claim to cover all the cases of the
techniques above, our approach is surprisingly powerful and effective. We
believe investigating and further understanding the reasons for its
effectiveness would be useful for better design of image editing with deep
learning.

\section{\name{}}
\label{sec:method}

In our setting, we are provided with a test image $\bx$ which we would
like to change  in a believable fashion with respect to a given
attribute. For example, the image could be a man without a beard, and
we would like to modify the image by adding facial hair while
preserving the man's identity. We further assume access to a set of
\emph{target} images \emph{with} the desired attribute
$\mathcal{S}^t=\left\{\bx_{1}^{t},...,\bx^{t}_{n}\right\}$ (e.g.,
\emph{men with facial hair}) and a set of \emph{source} images
\emph{without} the attribute
$\mathcal{S}^s=\left\{\bx_{1}^{s},...,\bx^{s}_{m}\right\}$ (e.g.,
\emph{men without facial hair}).
Further, we are provided with a pre-trained convnet trained on a sufficiently rich object categorization task---for example, the openly available VGG network~\cite{simonyan2014very} trained on ImageNet~\cite{ILSVRC15}. 
We can use this convnet to obtain a new representation of an image, which we denote as $\bx\rightarrow\phi(\bx)$. 
The vector $\phi(\bx)$ consists of concatenated activations of the convnet when applied to image $\bx$. We refer to it as the \emph{deep feature representation} of $\bx$.

\medskip
\noindent{\bf \name{}} can be summarized in four high-level
steps (illustrated in Figure~\ref{fig:overview}): 
\begin{packed_enum}
\item We map the images in the target and source sets \target and \source  into the deep feature representation through the pre-trained convnet $\phi$ (e.g., VGG-19 trained on ILSVRC2012).
\item We compute the mean feature values for each set of images, $\meantarget$  and $\meansource$,
 and define their difference as the \emph{attribute vector} 
\begin{equation}
	\bw=\bar{\phi}^t-\bar{\phi}^s.
\end{equation}
\item We map the test image $\bx$ to a point $\phi(\bx)$ in deep
feature space and move it along the attribute vector $\bw$, resulting in  $\phi(\bx)+\alpha \bw$. 
\item We can reconstruct the transformed output image $\bz$ by solving the reverse mapping into pixel space w.r.t. $\bz$
\begin{equation}
\intermediate = \phi(\bx)+\alpha  \bw.
\end{equation}
\end{packed_enum}
Although this procedure may appear deceptively simple, we show in Section~\ref{sec:results} that it can be surprisingly effective. In the following we will describe some important details to make the procedure work in practice. 

\medskip
\noindent{\bf Selecting \target and \source.}
\nameshort{} assumes that the attribute vector $\bw$ isolates the targeted transformation, i.e., it points towards the deep feature representation of image $\bx$ with the desired attribute change. 
If such an image $\bz$ was available (e.g., the same image of Mr. Robert Downey Jr. with beard), we could compute $\bw=\phi(\bz)-\phi(\bx)$ to isolate exactly the difference induced by the change in attribute. In the absence of the exact target image, we estimate $\bw$ through the target and source sets. It is therefore important that both sets are as similar as possible to our test image $\bx$ and there is no systematic attribute bias across the two data sets. If, for example, all target images in $\mathcal{S}^t$ were images of more senior people and source images in $\mathcal{S}^s$ of younger individuals, the vector $\bw$ would unintentionally capture the change involved in aging. Also, if the two sets are too different from the test image (e.g., a different race) the transformation would not look believable. 
To ensure sufficient similarity we restrict $\target$ and $\source$ to
the $K$ nearest neighbors of $\bx$. Let $\mathcal{N}^t_K$ denote the
$K$ nearest neighbors of $\target$ to $\phi(\bx)$; we define
\begin{equation}
 	\meantarget = \frac{1}{K}\sum_{\bx^t\in \mathcal{N}^t_K}\phi(\bx^{t}) \textrm { and }\meansource = \frac{1}{K}\sum_{\bx^s\in \mathcal{N}^s_K}\phi(\bx^{s}).
\end{equation} 
These neighbors can be selected in two ways,
depending on the amount of information available. When attribute
labels are available, we find the nearest images by counting the number of matching attributes (e.g., matching gender, race, age, hair color).
When attribute labels are unavailable, or as a second selection criterion, we take the nearest neighbors by cosine distance in deep feature space.

\medskip
\noindent{\bf Deep feature mapping.} There are many choices for a mapping into
deep feature space $\bx \rightarrow \phi(\bx)$.  We use the convolutional
layers of the normalized VGG-19 network pre-trained on ILSVRC2012, which has proven to be effective at
artistic style transfer~\cite{gatys2015neural}. The deep feature space must be
suitable for two very different tasks: (1) linear interpolation and (2)
reverse mapping back into pixel space. For the  interpolation, it is
advantageous to pick deep layers that are further along the linearization
process of deep convnets~\cite{bengio2013better}. In contrast, for the reverse
mapping, earlier layers capture more details of the image~\cite{mahendran15understanding}. The VGG network is divided into
five pooling regions (with increasing depth). As an effective compromise we
pick the first layers from the last three regions, \texttt{conv3\_1},
\texttt{conv4\_1} and \texttt{conv5\_1} layers (after ReLU activations), flattened and concatenated. 
As the pooling layers of VGG reduce the dimensionality of the input image, we \emph{increase} the image resolution of small images to be at least $200\times 200$ before applying $\phi$. 

\medskip
\noindent{\bf Image transformation.} Due to the ReLU activations used in most convnets (including VGG), all dimensions in $\phi(\bx)$ are non-negative and the vector is  sparse. 
As we average over $K$ images (instead of a single image as in~\cite{bengio2013better}), we expect $\bar{\phi}^t,\bar{\phi}^s$ to have very small components in most features. As the two data sets $\target$ and $\source$ only differ in the target attribute, features corresponding to visual aspects unrelated to this attribute will be averaged to very small values and approximately subtracted away in the vector $\bw$.

\medskip
\noindent{\bf Reverse mapping.} The final step of \nameshort{} is to reverse map the vector $\phi(\bx)+\alpha \bw$ back into pixel space to obtain an output image $\bz$. Intuitively, $\bz$ is an image that corresponds to $\phi(\bz)\approx\phi(\bx)+\alpha \bw$ when mapped into deep feature space.
Although no closed-form inverse function exists for the VGG mapping, we can obtain a color image by adopting the approach of~\cite{mahendran15understanding} and find  $\bz$ with gradient descent:
\begin{equation}
\bz=\argmin_{\bz} \frac{1}{2} \Vert(\phi(\bx)+\alpha\bw) - \phi(\bz)\Vert^2_{2} + \lambda_{V^{\beta}} R_{V^{\beta}}(\bz)\label{eq:under},
\end{equation} 
where $R_{V^{\beta}}$ is the Total Variation regularizer~\cite{mahendran15understanding} which encourages smooth transitions between neighboring pixels, 
\begin{equation}
R_{V^{\beta}}(\bz) \!=\! \sum_{i,j} \left( (z_{i,j+1} - z_{i,j})^2 + (z_{i+1,j} - z_{i,j})^2 \right)^{\frac{\beta}{2}}
\end{equation} 
Here, $z_{i,j}$ denotes the pixel in location $(i,j)$ in image $\bz$. 
Throughout our experiments, we set $\lambda_{V^{\beta}} = 0.001$ and $\beta=2$. 
We solve~(\ref{eq:under}) with the standard hill-climbing algorithm L-BFGS~\cite{liu1989limited}.

\label{sec:results}
\begin{figure*}[t]
\centerline{\includegraphics[width=\linewidth]{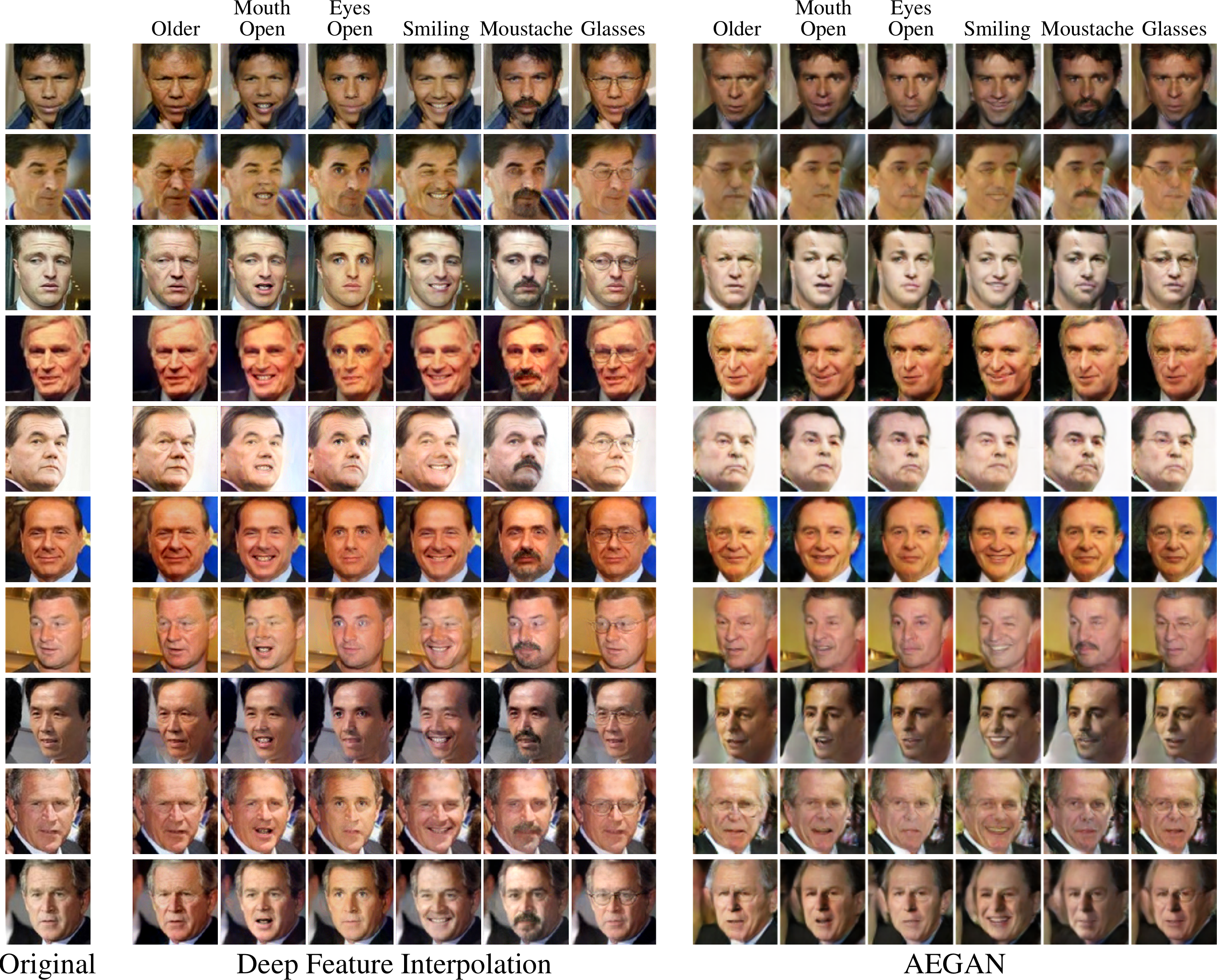}}
\caption{\textbf{(Zoom in for details.)} Adding different attributes
to the same person (random test images). \textbf{Left.} Original
image. \textbf{Middle.} \nameshort{}. \textbf{Right.} AEGAN. The goal
is to add the specified attribute while preserving the identity of the
original person.  For example, when adding a moustache to Ralf Schumacher
(3rd row) the hairstyle, forehead wrinkle, eyes looking to the right,
collar and background are all preserved by \nameshort{}. 
No foreground mask or human annotation was used to produce these
test results.}
\label{fig:attribute-matrix}
\vspace{-3ex}
\end{figure*}

\begin{figure*}[t]
\begin{center}
   \includegraphics[width=\linewidth]{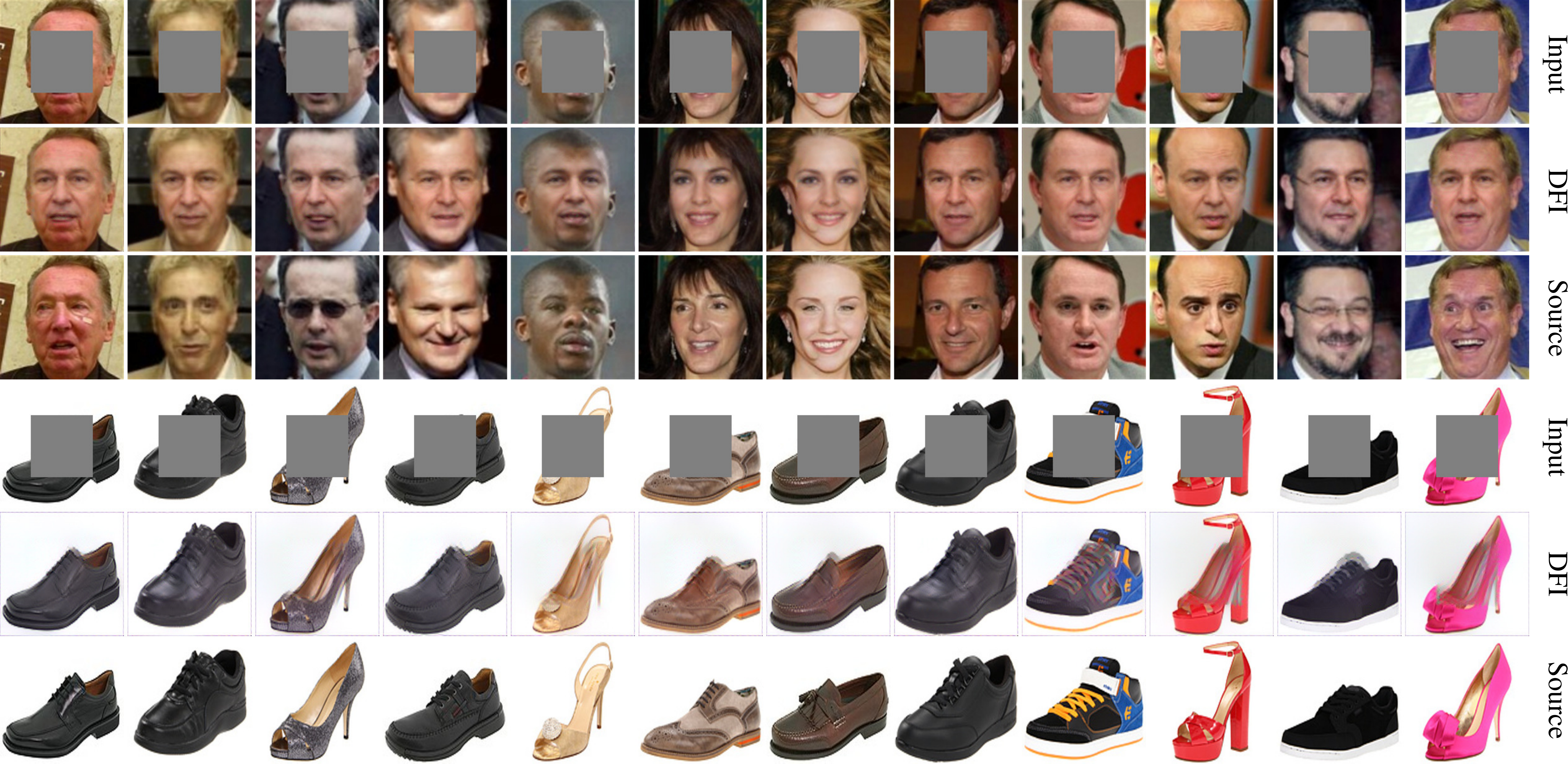}
\end{center}
\caption{\textbf{(Zoom in for details.)} Filling missing
regions. \textbf{Top.} LFW faces. \textbf{Bottom.} UT Zappos50k
shoes. Inpainting is an interpolation from masked to unmasked images. Given any dataset we
can create a source and target pair by simply masking out the missing
region. \nameshort{} uses $K\!=\!100$ such pairs derived from the nearest
neighbors (excluding test images) in feature space. The face results match wrinkles, skin tone,
gender and orientation (compare noses in 3rd and 4th images) but fail
to fill in eyeglasses (3rd and 11th images). The shoe results match
style and color but exhibit silhouette ghosting due to misalignment of
shapes. Supervised attributes were not used to produce these results. For the curious, we include the source image but we note that the goal is to produce a plausible region filling---not to reproduce the source.}
\label{fig:inpaint}
\vspace{-3ex}
\end{figure*}

\section{Experimental Results}

We evaluate \nameshort{} on a variety of tasks and data sets. 
For perfect reproducibility our code is available at \url{https://github.com/paulu/deepfeatinterp}.

\subsection{Changing Face Attributes}

We compare \nameshort{} to AEGAN~\cite{larsen2015autoencoding}, a
generative adversarial autoencoder, on several face attribute modification tasks. Similar to \nameshort{}, AEGAN also makes changes to faces by vector operations in a feature space.
We use the Labeled Faces in the Wild (LFW) data set, which contains 13,143 images of faces with predicted annotations for 73 different attributes (e.g., {\sc{sunglasses}}, {\sc{gender}}, {\sc{round face}}, {\sc{curly hair}}, {\sc{mustache}}, etc.). We use these annotations as attributes for our experiments. 
We chose six attributes for testing: \textsc{senior}, \textsc{mouth slightly open},
\textsc{eyes open}, \textsc{smiling}, \textsc{moustache} and \textsc{eyeglasses}. (The negative attributes are
\textsc{youth}, \textsc{mouth closed}, {\textsc{narrow eyes}}, \textsc{frowning}, \textsc{no beard}, \textsc{no eyewear}.) 
These attributes were chosen because 
it would be plausible for a single person
to be changed into having each of those attributes. Our test set consists of 38 images that did not have any of the six target attributes, were not \textsc{wearing hat}, had \textsc{mouth closed}, \textsc{no beard} and \textsc{no eyewear}. 
As LFW is highly gender imbalanced, we only used images of the more common gender, men, as target, source, and test images. 

Matching the approach of \cite{larsen2015autoencoding}, we align the face images and crop the outer pixels leaving a $100 \times 100$ face image, which we resize to $200\times 200$.
Target (source) collections are LFW images which have the positive (negative)
attributes. From each collection we take the $K=100$ nearest neighbors
(by number of matching attributes) to form \target and \source.

We empirically find that scaling $\bw$ by its mean squared feature activation makes the free parameter somewhat more consistent across multiple attribute transformations. If $d$ is the dimensionality of $\phi(\bx)$ and $pow$ is applied element-wise then we define
\begin{equation}
\alpha = \frac{\beta}{\frac{1}{d}pow(\bw,2)}.
\end{equation}
We set $\beta=0.4$.

Comparisons are shown in Figure~\ref{fig:attribute-matrix}. Looking down each
column, we expect each image to express the target attribute. Looking across
each row, we expect to see that the identity of the person is preserved.
Although AEGAN often produces the right attributes, it does not preserve identity as well as the much simpler \nameshort{}. 

\vspace{-2ex}
\paragraph{Perceptual Study.} Judgments of visual image changes are inherently subjective. To obtain an objective comparison between \nameshort{} and AEGAN we conducted a blind perceptual study with Amazon Mechanical Turk workers. 
We asked workers to pick
the image which best expresses the target attribute while preserving the
identity of the original face. This is a nuanced task so we required workers to
complete a tutorial before
participating in the study.  The task was a forced choice between AEGAN
and \nameshort{} (shown in random order) for six attribute changes on
38 test images. We collected an average of 29.6 judgments per image from
136 unique workers and found that \nameshort{} was preferred to AEGAN by a
ratio of 12:1. The least preferred transformation was Senior at 4.6:1 and
the most preferred was Eyeglasses at 38:1 (see Table~\ref{tab:face-study}).

\begin{table}
\begin{center}
\resizebox{\linewidth}{!}{
\begin{tabular}{c|c|c|c|c|c}
\multirow{2}{*}{older} & mouth & eyes & \multirow{2}{*}{smiling} & \multirow{2}{*}{moustache} & \multirow{2}{*}{glasses} \\
 & open & open & & \\
\hline
4.57 & 7.09 & 17.6 & 20.6 & 24.5 & 38.3
\end{tabular}
}
\end{center}
\caption{Perceptual study results. Each column shows the ratio at which workers preferred \nameshort{} to AEGAN on a specific attribute change (see Figure~\ref{fig:attribute-matrix} for images).}
\label{tab:face-study}
\vspace{-3ex}
\end{table}

\subsection{High Resolution Aging and Facial Hair}
\label{sec:highres}

One of the major benefits of DFI over many generative models is the ability to run on high resolution images. However, there are several challenges in presenting results on high resolution faces. 

First, we need a high-resolution dataset from which to select \source and \target. We collect a database of 100,000 high resolution face images from existing computer vision datasets (CelebA, MegaFace, and Helen) and Google image search~\cite{liu2015faceattributes, 2016mf2, le2012interactive}. We augment existing datasets, selecting only clear, unobstructed, front-facing high-resolution faces. This is different from many existing datasets which may have noisy and low-resolution images.

Next, we need to learn the attributes of the images present in the face dataset to properly select source and target images. Because a majority of images we collect do not have labels, we use face attribute classifiers developed using labeled data from LFW and CelebA.

Finally, the alignment of dataset images to the input image needs to be as close as possible, as artifacts that result from poor alignment are more obvious at higher resolutions. Instead of aligning our dataset as a preprocessing step, we use an off-the-shelf face alignment tool in DLIB \cite{kazemi2014one} to align images in \source and \target to the input image at test time.

We demonstrate results on editing megapixel faces for the tasks of aging and adding facial hair on three different faces. Due to the size of these images, selected results are shown in Figure~\ref{fig:high_res}. For full tables of results on these tasks, please see the supplementary materials.

\label{sec:results}
\begin{figure}
\centerline{\includegraphics[width=\linewidth]{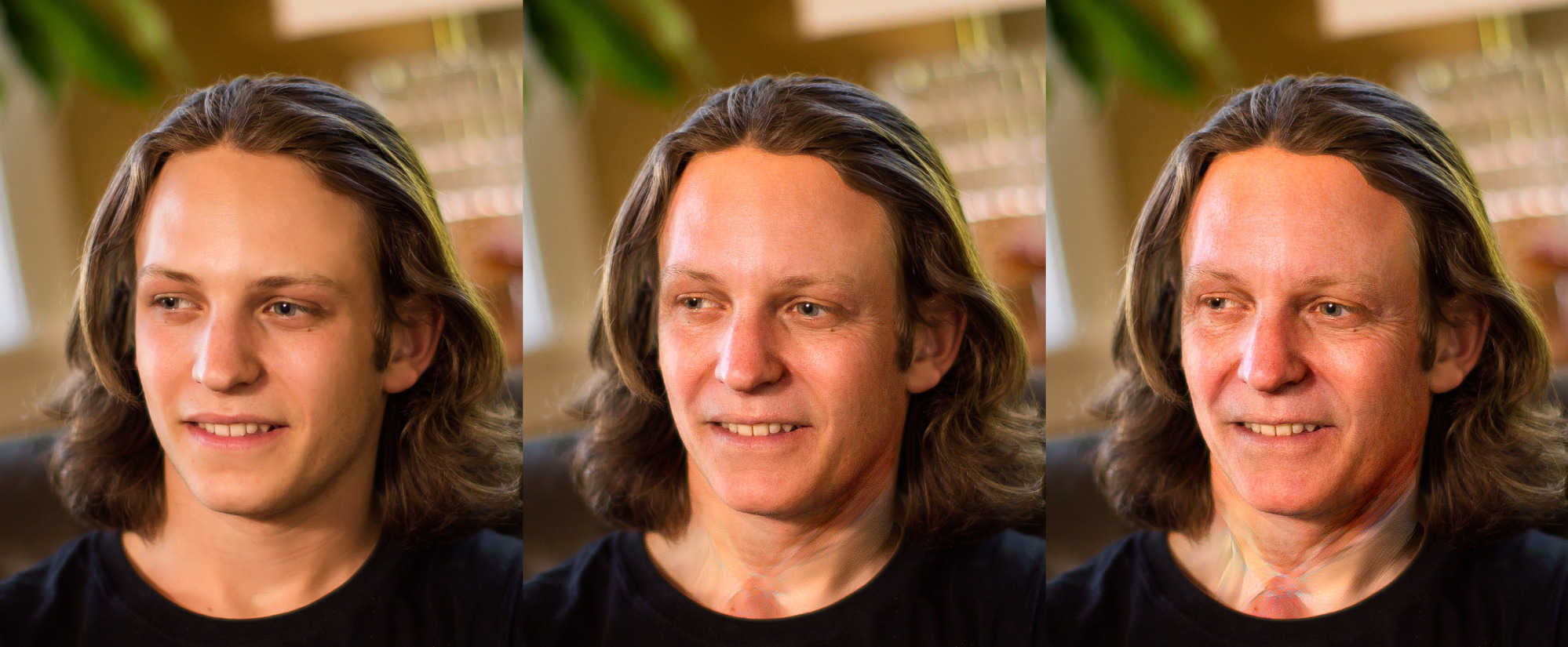}}
\centerline{\includegraphics[width=\linewidth]{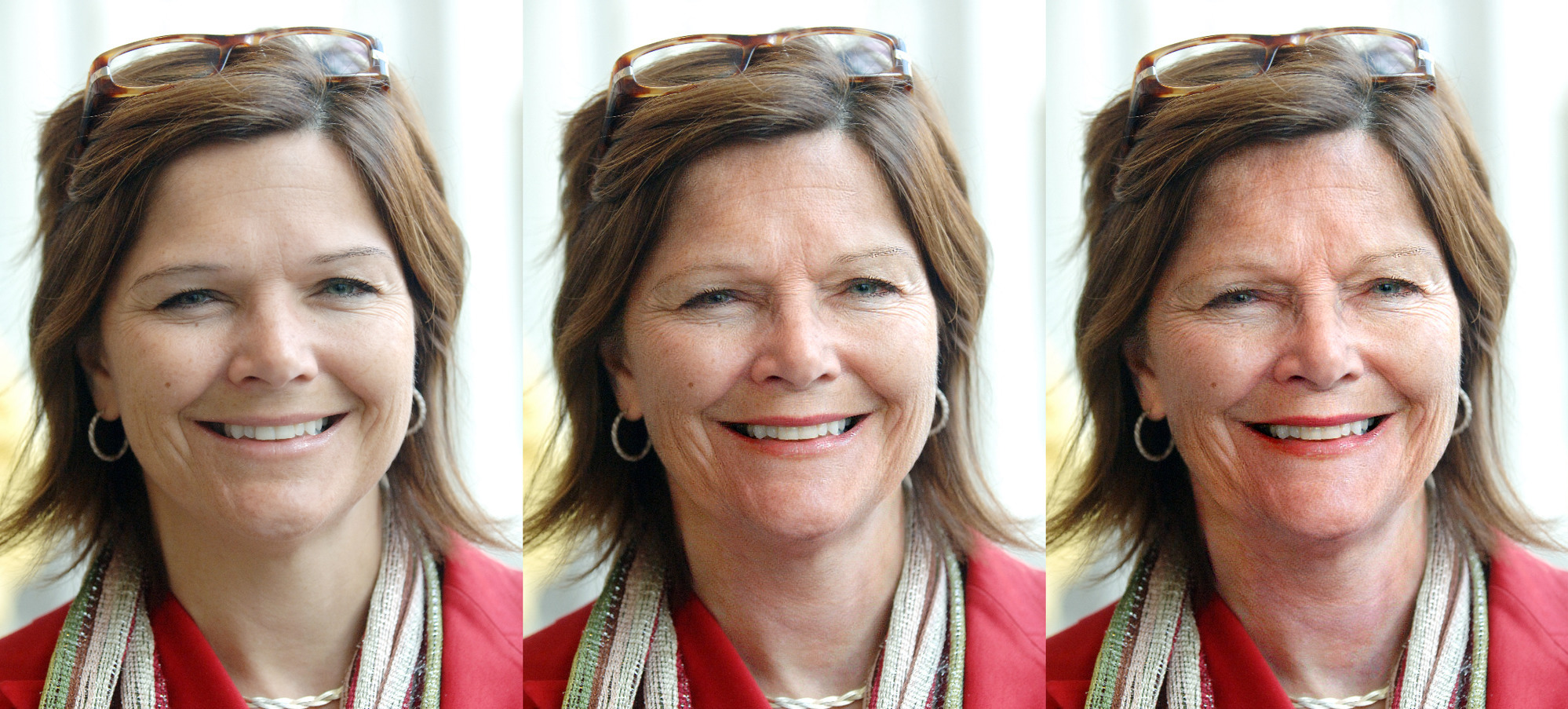}}
\centerline{\includegraphics[width=\linewidth]{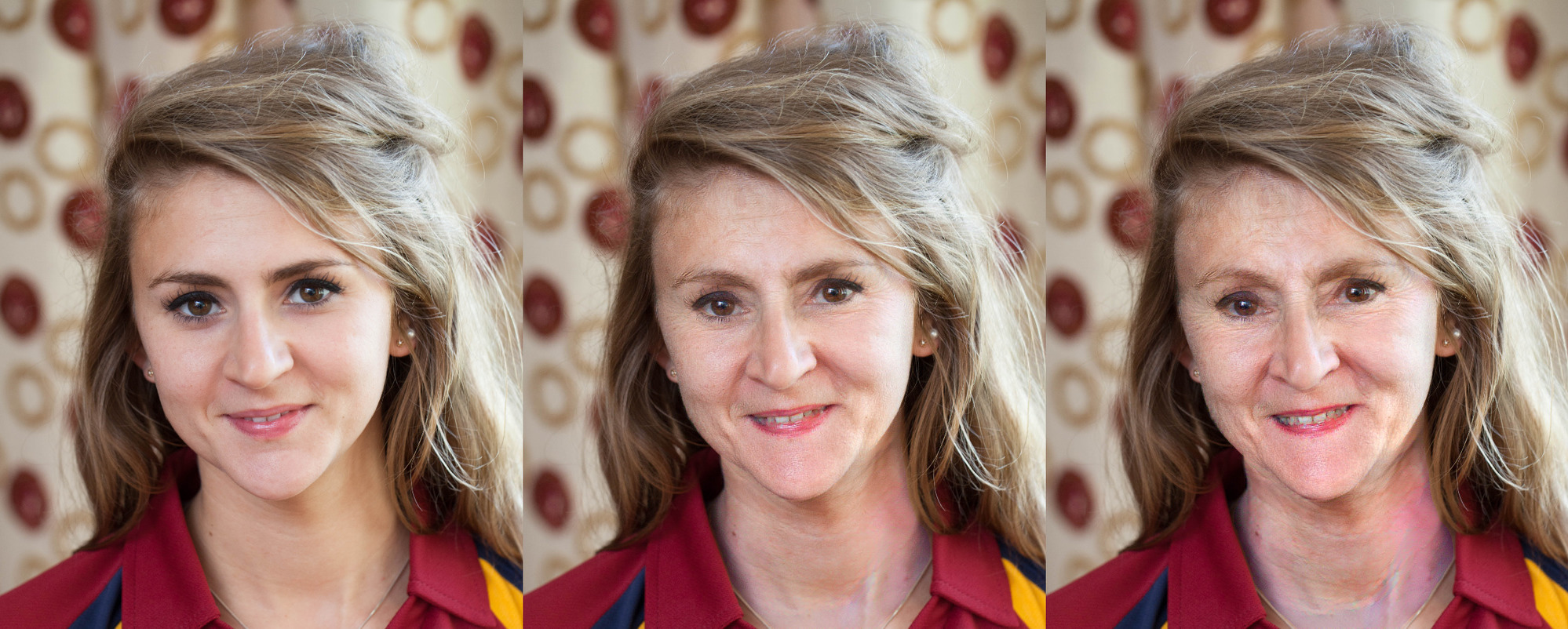}}
\centerline{\includegraphics[width=\linewidth]{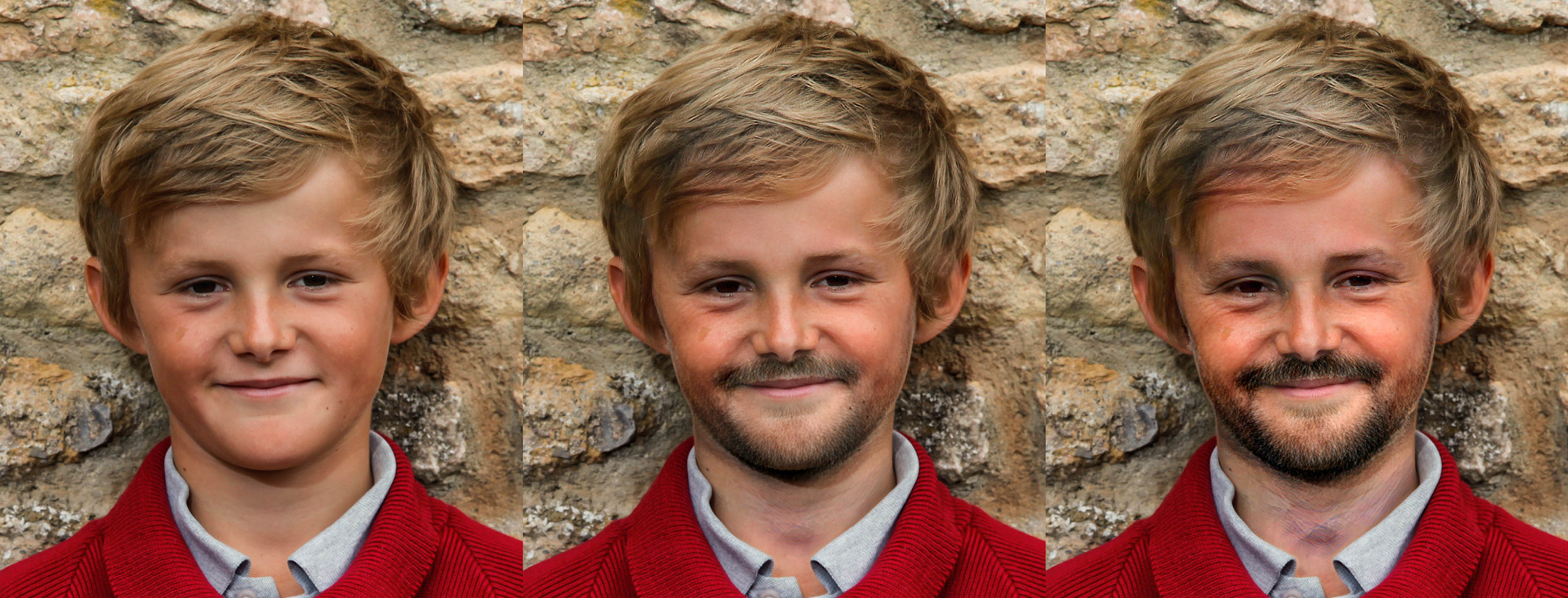}}
\centerline{\includegraphics[width=\linewidth]{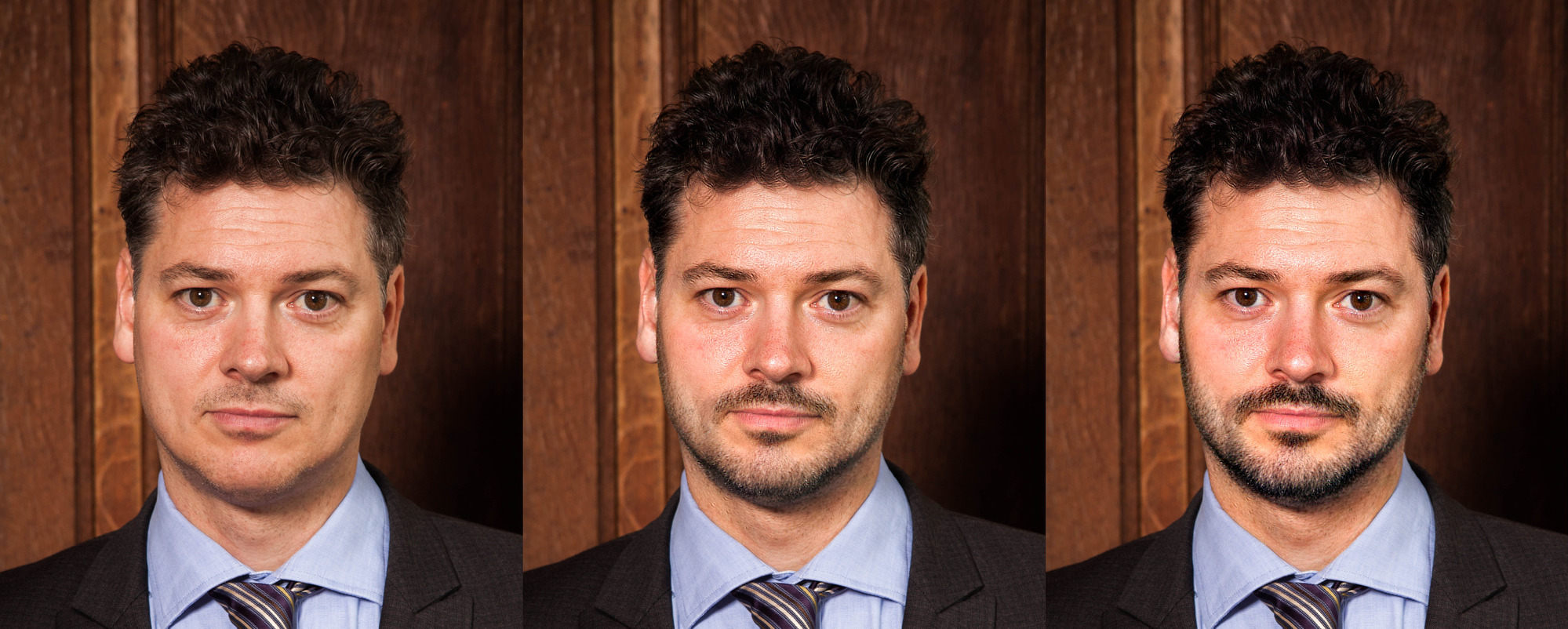}}
\centerline{\includegraphics[width=\linewidth]{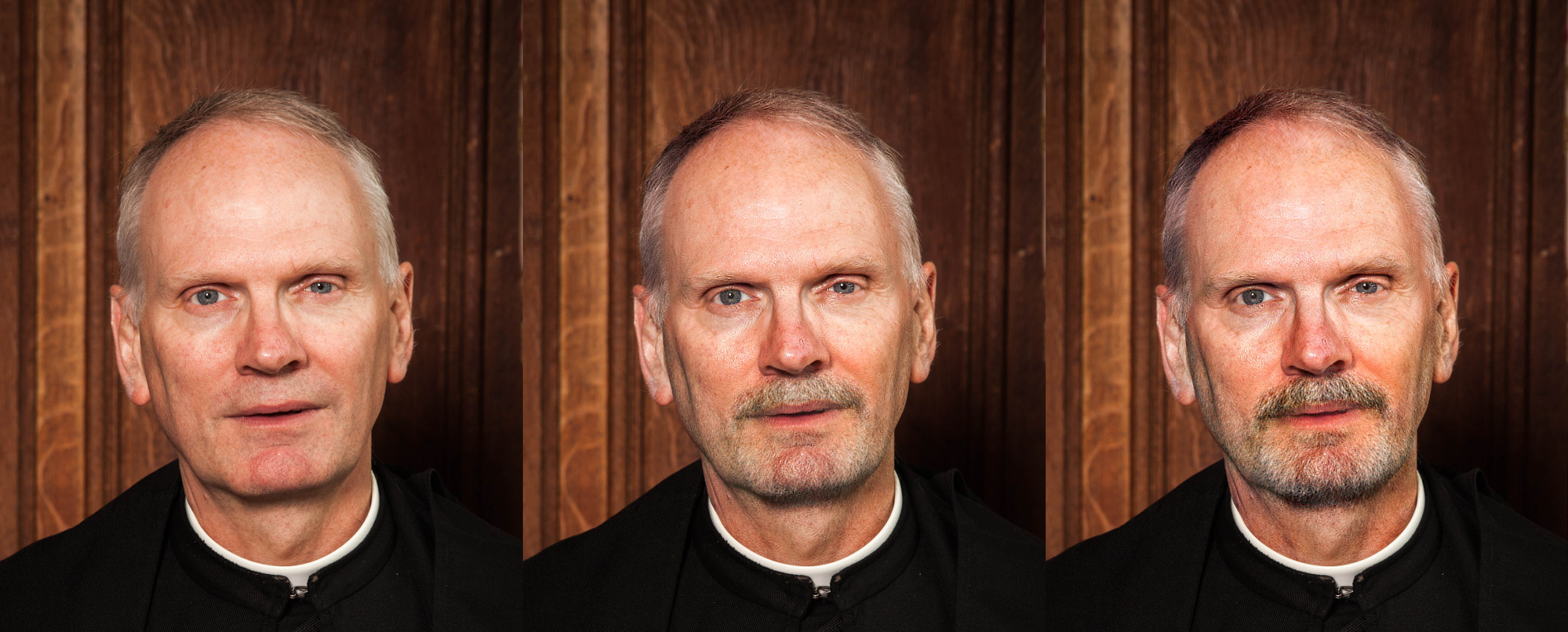}}
\caption{\textbf{(Zoom in for details.)} Editing megapixel faces.
\textbf{First column.} Original image. \textbf{Right columns.} The top 3 rows show aging ($\beta = \{0.15,0.25\}$) and the bottom 3 rows show the addition of facial hair ($\beta = \{0.21,0.31\}$).
High resolution images are challenging since artifacts are more perceivable. We find \nameshort{} to be effective on the aging and addition of facial hair tasks.
}
\label{fig:high_res}
\vspace{-3ex}
\end{figure}

\subsection{Inpainting Without Attributes}

Inpainting fills missing regions of an image with plausible pixel
values. There can be multiple correct answers. Inpainting is hard
when the missing regions are large (see Figure~\ref{fig:inpaint} for
our test masks). Since attributes cannot be predicted (e.g., eye color when both eyes are missing) we use distance in feature space to select the nearest neighbors.

Inpainting may seem like a very different task from changing face attributes,
but it is actually a straightforward application of \nameshort{}. All we need are source
and target pairs which differ only in the missing regions. Such pairs
can be generated for any dataset by taking an image and masking out
the same regions that are missing in the test image. The images with mask become the source set and those without the target set.  We then find the $K\!=\!100$ nearest neighbors in the masked dataset (excluding test images) by cosine distance in \mbox{VGG-19} pool5 feature space. We experiment on two datasets: all of LFW (13,143 images, including male and female images) and the Shoes subset of UT Zappos50k (29,771 images)~\cite{fine-grained,pathak2016context}. For each dataset
we find a single $\beta$ that works well ($1.6$ for LFW and $2.8$
for UT Zappos50k).

\begin{figure}[t]
\begin{center}
   \includegraphics[width=0.85\linewidth]{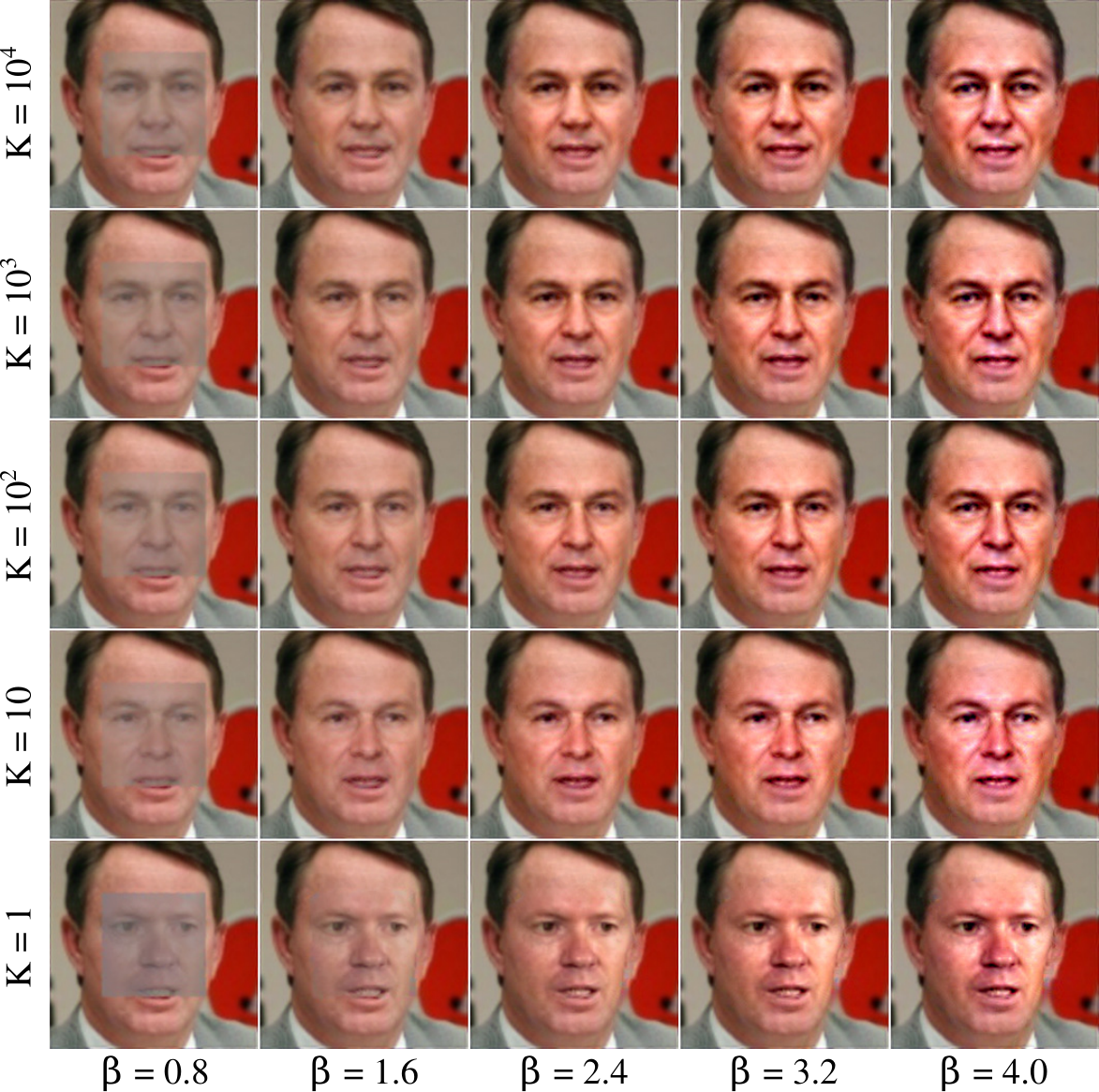}
\end{center}
\caption{Inpainting and varying the free parameters. \textbf{Rows:} $K$, the number of nearest neighbors. \textbf{Columns:} $\beta$, higher values correspond to a larger perturbation in feature space. When $K$ is too small the generated pixels do not fit the existing pixels as well (the nose, eyes and cheeks do not match the age and skin tone of the unmasked regions). When $K$ is too large a difference of means fails to capture the discrepancy between the distributions (two noses are synthesized). When $\beta$ is too small or too large the generated pixels look unnatural. We use $K=100$ and $\beta=1.6$.}
\label{fig:vary_params}
\end{figure}

We show our results in Figure~\ref{fig:inpaint} on 12 test
images (more in supplemental) which match those used by
disCVAE~\cite{yan2015attribute2image} (see Figure 6 of their
paper). 
Qualitatively we observe that the \nameshort{} results are plausible. The filled face regions match skin tone,
wrinkles, gender, and pose. The filled shoe regions match color
and shoe style. However, \nameshort{} failed to produce eyeglasses when stems are visible in the input and some shoes exhibit ghosting since the dataset is not perfectly aligned. \nameshort{} performs well when the face is missing (i.e., the central portion of each image) but we found it performs worse than disCVAE when half of the image is missing (Figure~\ref{fig:inpaint_half}). Overall, \nameshort{} works surprisingly well on these inpainting tasks. The results are  particularly impressive considering that, in contrast to disCVAE, it does not require attributes to describe the missing regions.

\begin{figure*}[t!]
\begin{center}
   \includegraphics[width=0.95\linewidth]{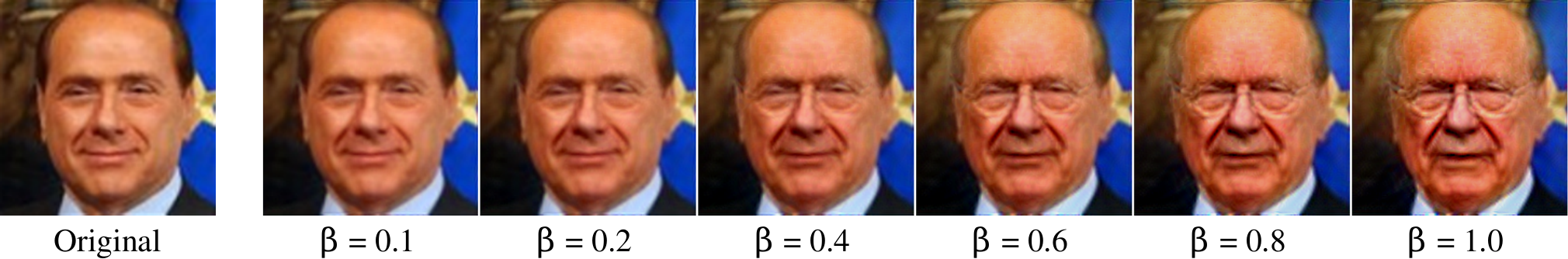}
\end{center}
\vspace{-1.5ex}
\caption{Morphing a face to make it appear older. The transformation becomes more pronounced as the value of $\beta$ increases.} 
\label{fig:age_morph}
\vspace{-1.5ex}
\end{figure*}

\subsection{Varying the free parameters} 
Figure~\ref{fig:vary_params} illustrates the effect of changing $\beta$ (strength of transformation) and $K$ (size of source/target sets). As $\beta$ increases, task-related visual elements change more noticeably (Figure~\ref{fig:age_morph}). If $\beta$ is low then ghosting can appear. If $\beta$ is too large then the transformed image may jump to a point in feature space which leads to an unnatural reconstruction. $K$ controls the variety of images in the source and target sets. A lack of variety can create artifacts where changed pixels do not match nearby unchanged pixels (e.g., see the lower lip, last row of Figure~\ref{fig:vary_params}). However, too much variety can cause \target and \source to contain distinct subclasses and the set mean may describe something unnatural (e.g., in the first row of Figure~\ref{fig:vary_params} the nose has two tips, reflecting the presence of left-facing and right-facing subclasses). In practice, we pick an $\beta$ and $K$ which work well for a variety of images and tasks rather than choosing per-case.

\section{Discussion}

In the previous section we have shown that \name{} is surprisingly effective on several image transformation tasks. This is very promising and may have implications for future work in the area of automated image transformations. However, \nameshort{} also has clear limitations and requirements on the data. We first clarify some of the aspects of \nameshort{} and then focus on some general observations.

\paragraph{Image alignment} is a necessary requirement for \nameshort{} to work. 
We use the difference of means to cancel out the contributions of
convolutional features that are unrelated to the attribute we wish to change, particularly when this attribute is centered in a specific location (adding a mustache, opening eyes, adding a smile, etc). For example, when adding a mustache, all target images contain a mustache and therefore the convolutional features with the mustache in their receptive field will not average out to zero. While max-pooling affords us some degree of translation invariance, this reasoning breaks down if mustaches appear in highly varied locations around the image, because no specific subset of convolutional features will then correspond to ``mustache features''.
Image alignment is a limitation but not for faces, an important class of images. As shown in Section~\ref{sec:highres}, existing face alignment tools are sufficient for \nameshort{}.

\paragraph{Time and space complexity.}
A significant strength of \nameshort{} is that it is very lean.
The biggest resource footprint is GPU memory for the convolutional layers of VGG-19 (the large fully-connected layers are not needed). A $1280 \times 960$ image requires 4 GB and takes 5 minutes to reconstruct. A $200\times200$ image takes 20s to process. The time and space complexity are linear. In comparison, many generative models only demonstrate $64 \times 64$ images. Although \nameshort{} does not require the training of a specialized architecture, it is also fair to say that during test-time it is significantly slower than a trained model (which, typically, needs sub-seconds) As future work it may be possible to incorporate techniques from real-time style-transfer~\cite{sadeghi2015visalogy} to speed-up \nameshort{} in practice.

\paragraph{\nameshort{}'s simplicity.}
Although there exists work on high-resolution style transfer~\cite{gatys2015neural,mahendran15understanding,sadeghi2015visalogy}, 
to our knowledge, \nameshort{} is the first algorithm to enable automated high resolution content transformations. The simple mechanisms of \nameshort{} may inspire more sophisticated follow-up work on scaling up current generative architectures to higher resolutions, which may unlock a wide range of new applications and use cases.

\paragraph{Generative vs. Discriminative networks.}
To our knowledge, this work is the first cross-architectural comparison of an AE against a method that uses features from a discriminatively trained network. To our great surprise, it appears that a discriminative model has a latent space as good as an AE model at editing content. A possible explanation is that the AE architecture could organize a better latent space if it were trained on a more complex dataset. AE are typically trained on small datasets with very little variety compared to the size and richness of recognition datasets. The richness of ImageNet seems to be an important factor: in early experiments we found that the convolutional feature spaces of VGG-19 outperformed those of VGG-Face on face attribute change tasks. 

\begin{figure}[t]
\begin{center}
   \includegraphics[width=0.85\linewidth]{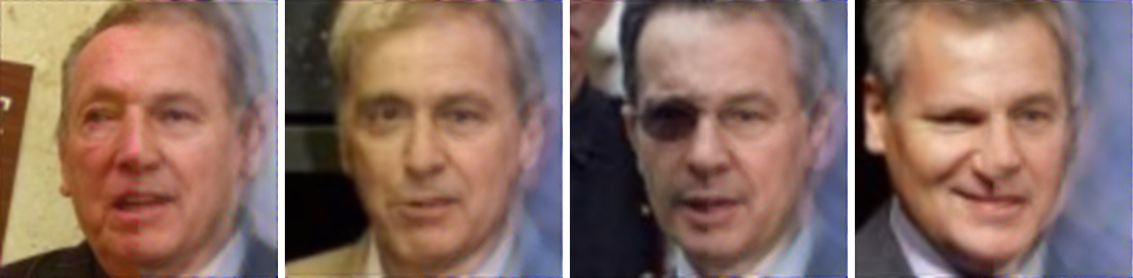}
\end{center}
\caption{Example of a hard task for \nameshort{}: inpainting an image with the right half missing.}
\label{fig:inpaint_half}
\end{figure}

\paragraph{Linear interpolation as a baseline.}
Linear interpolation in a pre-trained feature space can serve as a first test for determining if a task is interesting: problems that can easily be solved by \nameshort{} are unlikely to require the complex machinery of generative networks.
Generative models can be much more powerful than
linear interpolation, but the current problems (in particular, face attribute editing) which are used to showcase generative approaches are too simple. Indeed, we do find many problems where generative models outperform \nameshort{}. In the case of inpainting we find \nameshort{} to be lacking when the masked region
is half the image (Figure~\ref{fig:inpaint_half}). \nameshort{}
is also incapable of shape~\cite{zhu2016generative} or
rotation~\cite{reed2015deep} transformations since those tasks require
aligned data. Finding more of these difficult tasks where generative models outshine \nameshort{} would help us better evaluate generative models. 
We
propose \nameshort{} to be the linear interpolation baseline because it
is very easy to compute, it will scale to future high-resolution models,
it does not require supervised attributes, and it can be applied to
nearly any aligned class-changing problems.

\section{Conclusion}
We have introduced \nameshort{} which interpolates in a pre-trained feature space to achieve a wide range of image transformations like aging, adding facial hair and inpainting.
Overall, \nameshort{} performs surprisingly well given the method's simplicity. It is able to produce high quality images over a variety of tasks, in many cases of higher quality than existing state-of-the-art methods. This suggests that, given the ease with which \nameshort{} can be implemented, it should serve as a highly competitive baseline for certain types of image transformations on aligned data. Given the performance of \nameshort{}, we hope that this spurs future research into image transformation methods that outperform this approach.

\paragraph{Acknowledgments}
JRG, and KQW are supported in part by the III-1618134, III-1526012, and IIS-1149882 grants from the National Science Foundation, as well as the Bill and Melinda Gates Foundation. PU and KB are supported by the National Science Foundation (grants IIS-1617861, IIS-1011919, IIS-1161645, IIS-1149393), and by a Google Faculty Research Award. NS is supported by the National Science Foundation grant IIS-1149393.

{\small
\bibliographystyle{ieee}
\bibliography{all}
}

\end{document}